\documentclass{article}

\usepackage{amsmath,epsfig}
\usepackage[preprint]{spconf}
\usepackage{amssymb}
\usepackage{bm}
\usepackage{cite}
\usepackage{url}
\usepackage{booktabs}
\usepackage{multirow}
\usepackage{makecell}
\usepackage{xcolor}
\usepackage{colortbl}
\usepackage{algorithm,algpseudocode}
\usepackage{fancyhdr}
\usepackage{flushend}
\usepackage{comment}

\algrenewcommand\alglinenumber[1]{\footnotesize #1:}

\AtBeginDocument{\colorlet{defaultrulecolor}{.}}

\DeclareMathOperator*{\argmin}{arg\,min}

\begin{document}

\ninept

\title{LIGHTWEIGHT COMPRESSION OF NEURAL NETWORK FEATURE TENSORS FOR COLLABORATIVE INTELLIGENCE}
\address{}
\name{Robert A. Cohen, Hyomin Choi, and Ivan V. Baji\'{c}}
\address{School of Engineering Science, Simon Fraser University, Burnaby, BC, Canada}

\maketitle

\fancypagestyle{firstpage}
{
    \vspace{-3\baselineskip}
    \fancyhf{}
    \fancyhead[C]{Copyright \copyright 2020 IEEE. Personal use of this material is permitted. Permission from IEEE must be obtained for all other uses, in any current or future media, including reprinting/republishing this material for advertising or promotional purposes, creating new collective works, for resale or redistribution to servers or lists, or reuse of any copyrighted component of this work in other works.}
    \setlength{\headheight}{48pt}
    \renewcommand{\headrulewidth}{0.0pt}
}

\thispagestyle{firstpage}
\renewcommand{\headrulewidth}{0.0pt}

\begin{abstract}
In collaborative intelligence applications, part of a deep neural network (DNN) is deployed on a relatively low-complexity device such as a mobile phone or edge device, and the remainder of the DNN is processed where more computing resources are available, such as in the cloud. This paper presents a novel lightweight compression technique designed specifically to code the activations of a split DNN layer, while having a low complexity suitable for edge devices and not requiring any retraining. We also present a modified entropy-constrained quantizer design algorithm optimized for clipped activations. When applied to popular object-detection and classification DNNs, we were able to compress the 32-bit floating point activations down to 0.6 to 0.8 bits, while keeping the loss in accuracy to less than 1\%. When compared to HEVC, we found that the lightweight codec consistently provided better inference accuracy, by up to 1.3\%. The performance and simplicity of this lightweight compression technique makes it an attractive option for coding a layer's activations in split neural networks for edge/cloud applications.
\end{abstract}

\section{Introduction}
\label{sec:intro}

With recent advances in machine learning and their efficient implementations, we are beginning to see an increasingly widespread use of deep neural networks (DNNs) on small or lightweight platforms such as mobile devices, edge computing systems, and embedded platforms~\cite{Chen2019,Lane2018,Tan_2019_CVPR}. For DNNs that are too large or complex to realize entirely within a mobile or edge device, a collaborative intelligence~\cite{eshratifar2019towards} approach can be used to split the DNN between the device and the cloud. The data output at the split location is signaled to the remainder of the DNN on the cloud. Determining the ideal place to split the network can depend upon both the application and the size of the activations or feature tensors that need to be signaled~\cite{Kang2017}. The feature tensors inside a DNN may be greatly expanded in terms of size and redundancy as compared to the data originally input to the network. Therefore, data reduction or compression methods are needed in order to reduce the bandwidth of the data to be transferred.

Existing approaches to reducing the data bandwidth between layers typically include quantization, data reorganization, pruning, and sometimes the addition of small neural networks to reduce the feature tensor dimensions. Quantization within DNNs is currently a very active area of research, not only for collaborative intelligence, but also for reducing overall complexity and memory requirements. In~\cite{Mishra2017_WRPN}, ResNet\nobreakdash-34 was modified to have more filters in each layer, and the weights and activations were quantized to 2 and 4 bits, respectively, while maintaining the accuracy of the unmodified network. In~\cite{Banner2018_8BitTraining}, weights, activations, and some of the gradient and back-propagation computations were quantized to 8 bits, along with a scaling modification made to batch normalization. In~\cite{Choi2019_2bit}, networks were trained with weights and activations quantized to two bits. Binary neural networks~\cite{Hubara2016_BNN,Rastegari2016_XNOR} use only one-bit weights and activations during inference. For collaborative intelligence applications, a convolutional layer can be inserted after the split layer to significantly reduce the dimensions of the feature tensor, followed by 8-bit quantization~\cite{eshratifar2019towards}. These methods, however, require end-to-end (re)training to obtain the final network weights and parameters. Given the complexity of the training process and availability of pre-trained floating-point weights for state of the art DNNs, methods for post-training quantization have been of great interest as well. Several post-training quantization techniques~\cite{Zhao2019_Quantization,banner2018_ACIQ,Banner2019_4bit,Krishnamoorthi2018_QuantizingDC} reduce the losses caused by quantization while quantizing weights and/or activations down to between 4 (or an average of 4) and 8 bits, and sometimes lower. Another method to reduce the data size is to compress tiled picture(s) of the activation tensors at the split layer using conventional image or video codecs~\cite{dfc_for_collab_object_detection, Choi2018NearLosslessDF, eshratifar2019bottlenet, eshratifar2019towards,Choi_BaF_2020}.

In this paper, we present a lightweight compression method that is well-suited for coding the output of a split DNN in edge-based devices. This method uses simple and very coarse scalar quantization along with clipping, binarization, and entropy coding to compress the activations. No retraining of network weights is needed. It is also capable of coding to a wide range of bit-rates. We also present comparisons to compression using the HEVC screen content coding extension~\cite{hevc_scc}, by tiling the activation channels into pictures as was done in~\cite{Choi2018NearLosslessDF}.

In Section~\ref{sec:lightweight}, we present the lightweight codec along with an examination of the effects of clipping. Section~\ref{sec:experiments} presents experimental results, followed by conclusions in Section~\ref{sec:conclusions}.

\section{Lightweight compression of tensors}
\label{sec:lightweight}
\begin{figure}[tb]
    \centering
    \includegraphics[width=0.48\textwidth,viewport=2.391047 225.827993 553.571983 538.217984,clip]{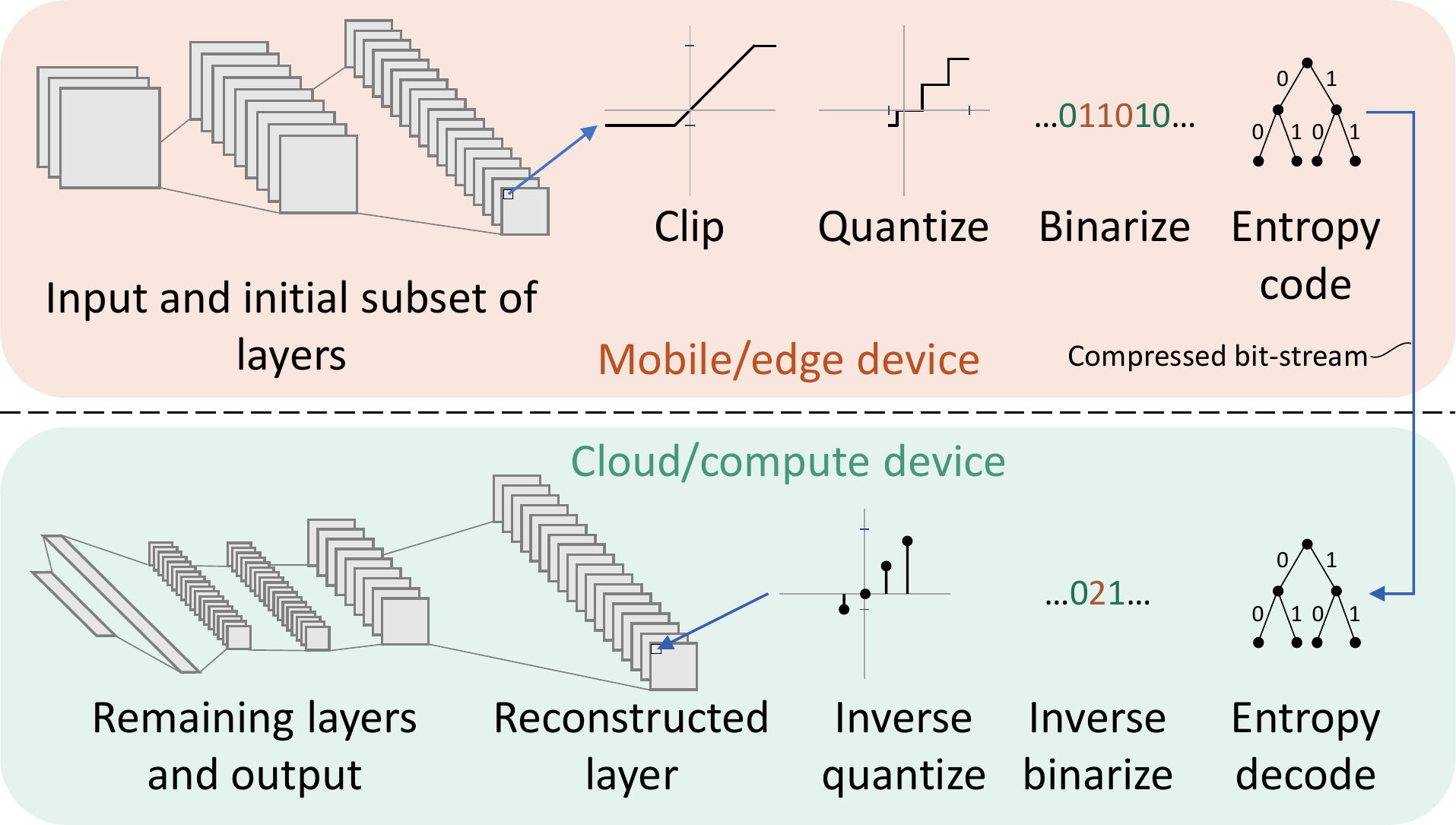}
    \caption{Lightweight compression system overview
    }
    \label{fig:system}
\end{figure}

Fig.~\ref{fig:system} illustrates how the lightweight compression method can be used in a collaborative intelligence application. Layers in a DNN include operations such as convolutions, batch normalization operations, and activation functions. If the first several layers of a DNN are performed for example on a mobile or edge device, we would like to compress the outputs of the activation functions (feature tensors) for transmission to the platform that is performing the remaining layers of the DNN. Because the compression will be performed on an edge device, we would like to keep the complexity of the compression method relatively low, while not significantly compromising the accuracy of the DNN model. For the lightweight compression presented here, we rely on relatively simple operations such as clipping and very coarse scalar quantization to a few representative values. To further compress the data, we convert the quantized symbols to a binarized representation for passing to an entropy coder. The compressed bitstream is sent to the cloud or another computing platform, where it is decoded and converted to a reconstructed feature tensor, which is then processed by the remaining layers of the DNN. The net effect of this process on the DNN computations is that the output of one layer is clipped and quantized; namely the layer that will be transmitted.

For post-training quantization of activations for applications such as object detection and classification, it is well known that uniformly quantizing
32-bit floating point activation values to 8 bits generates a small or negligible reduction in the network's accuracy~\cite{Vanhoucke2011}. This behavior also holds true for collaborative intelligence applications with quantization applied to where the network is split. The network's performance degrades significantly, however, when quantizing without refinements to 4 or fewer bits. For example, the Top\nobreakdash-1 classification accuracy of ResNet-50~\cite{He2015DeepRL} on the ImageNet ILSVRC2012~\cite{imagenet2015} validation data set is 75.8\%, but if we cut the network at the output of layer 21 and uniformly quantize each activation value to 8 levels (3 bits), then the accuracy becomes 59.7\%; a drop of 16.1\%.
By clipping the tensor elements prior to quantization, it is possible to eliminate this loss, without requiring any retraining of the DNN's weights. In this section, we first examine the effects of clipping, and then we present the lightweight compression technique which makes use of an entropy-constrained quantizer design process tailored for clipped activations. 

\pagestyle{empty}

\subsection{Effects of clipping}
\label{subsec:clipping}
\begin{figure}[ht]
    \centering
    \includegraphics[width=0.48\textwidth,viewport=7.740000 7.794000 701.243979 413.081987,clip]{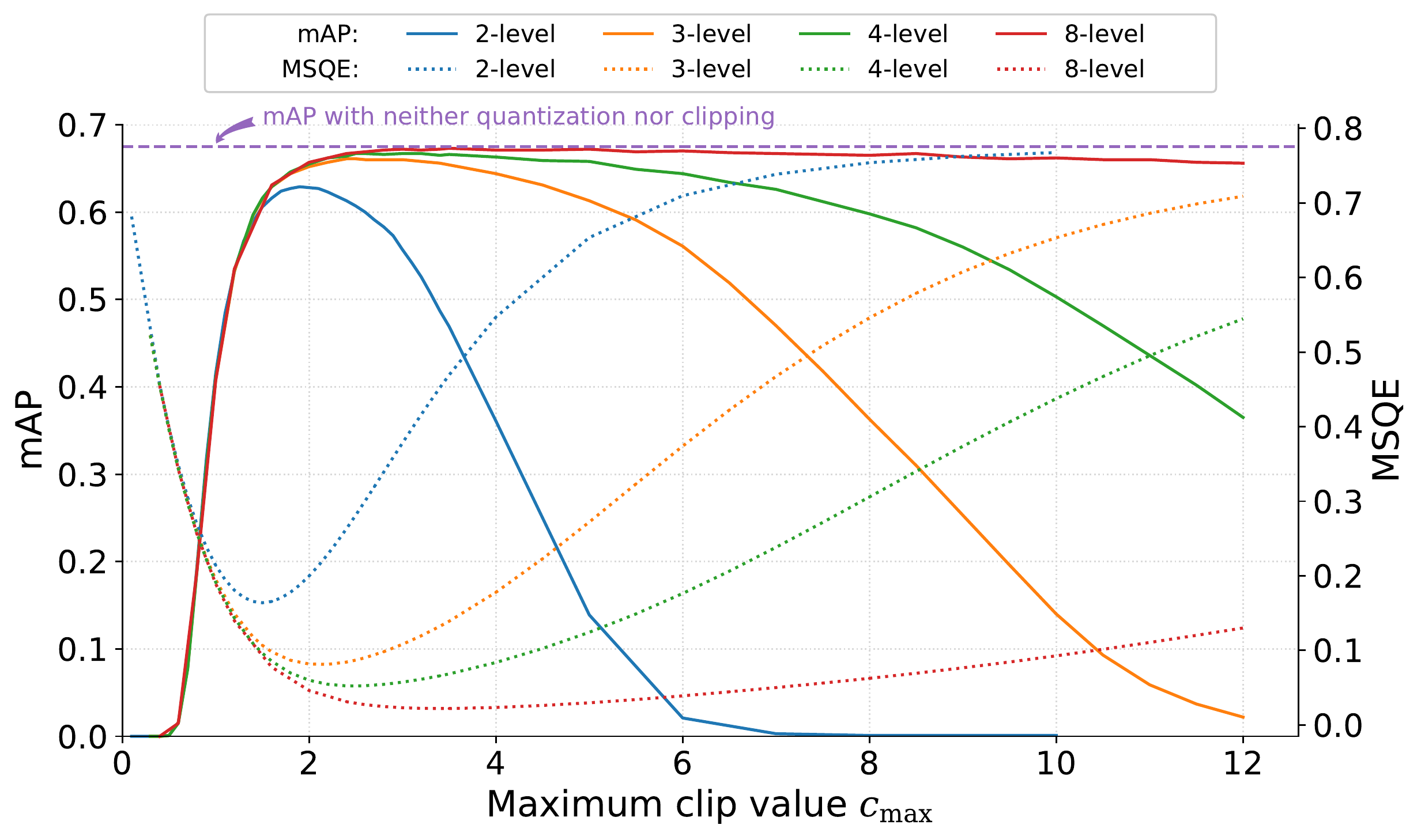}
    \caption{Effects of clipping layer 12 activations in YOLOv3}
    \label{fig:clip_range}
\end{figure}

Fig.~\ref{fig:clip_range} shows the effects that clipping and coarse quantization have on the mean Average Precision (mAP) of the YOLOv3~\cite{Redmon2018_yolov3} object detection network when run on the COCO 2017~\cite{COCO} validation data set (IoU = 0.5). Here, the activations at the output of layer 12 are clipped (clamped) to be between $c_\mathrm{min} = 0.0$ and $c_\mathrm{max}$. Each clipped activation value, denoted as $x_\mathrm{clp}$, is processed by an $N$-level quantizer as follows:
\begin{equation}
\label{eq:clipquant}
    Q(x_\mathrm{clp}) =  \mathrm{round}\left(\frac{x_\mathrm{clp} - {c}_\mathrm{min}}{{c}_\mathrm{max} - {c}_\mathrm{min}}
    \cdot (N-1)\right) \, ,
\end{equation}
where $\mathrm{round(\cdot)}$ rounds away from zero for halfway cases. Note that unlike related literature that focuses on reduced bit-depth architectures, our $N$ does not need to be a power of two, as the purpose of quantization is for compression and subsequent transmission or storage in a bit-stream. The mean-square quantization error (MSQE) computed between an unmodified activation $x$ and the inverse-quantized activation is also shown in Fig.~\ref{fig:clip_range}.

For 3-bit quantization, i.e. 8-level quantization, peak performance is achieved over a range of ${c}_\mathrm{max}$ values between roughly 3.0 and 5.0. The performance degrades when clipping extends further. As the number of quantization levels is decreased, the optimal ${c}_\mathrm{max}$ decreases, as does the range of ${c}_\mathrm{max}$ values that achieves peak performance. With 1-bit (2-level) quantization, the optimal range is quite narrow. When the quantization is not extremely coarse, e.g. 8-level (3-bit) or higher, the minimum MSQE generally coincides with the peak mAP performance. Earlier works, e.g.~\cite{banner2018_ACIQ,Banner2019_4bit} have leveraged this behavior to model the quantization error in order to select the optimal clipping range for all activations in a DNN. However, it is evident from these prior works that deviations from the models occur with extremely coarse quantization, e.g. 2-bit (4-level) and below, and when the distributions of the activations are non-Gaussian. We can see in Fig.~\ref{fig:clip_range} that the optimal ${c}_\mathrm{max}$ for 1-bit quantization is approximately 2.0, whereas the minimum MSQE occurs near
${c}_\mathrm{max}=1.75$.
Thus, choosing ${c}_\mathrm{max}$ based on minimizing MSQE can result in a potential loss in mAP of several percent when $N$ is small. For the experiments presented later in this paper, we empirically select the clipping ranges in order to determine the best achievable network accuracy when compression is used.

\subsection{Modified entropy-constrained quantization for clipped activations}
Although a uniform quantizer is simple, it is not optimal for signals that are not uniformly distributed. Moreover, because there is a trade-off between network accuracy and bit-rate when compressing activations, we would like be able to compress the tensors over a range of rates or file sizes. Entropy-constrained quantization~\cite{Chou1989} and rate-distortion optimization~\cite{Sullivan1998} are well-known methods for compressing data subject to minimizing a Lagrangian cost function $J = D + \lambda R$, where $D$ is a distortion metric, $R$ is a rate or size of the representation, and $\lambda$ is a scalar. Thus, we can easily obtain optimal quantizers in the mean-squared sense over a range of rates by using an entropy-constrained design process. However, we showed earlier that the accuracy of a DNN is quite sensitive to the clipping range of a layer's activation when quantized to a very low number of levels. The representative level for each bin of an $\ell^2$-norm optimized quantizer corresponds to the centroid of the data quantized to that bin. For example, suppose a one-bit quantizer divides $[0.0,2.0]$ into two bins, $[0.0,1.0]$ and $[1.0,2.0]$, and the representative values are 0.3 and 1.5, respectively. If a layer's activations are clipped to $[0.0,2.0]$, quantized, and transmitted, then the receiver would reconstruct data having only the values $\{0.3,1.5\}$, which span a range much smaller than the initial clipping range of $[0.0,2.0]$. To address this problem, we present a modified entropy-constrained quantizer design process to pin the reconstruction levels of the outermost bins to ${c}_\mathrm{min}$ and ${c}_\mathrm{max}$, to ensure that the reconstructed activations span the full clipping range. The reconstruction values for the interior bins and the threshold values between bins are not pinned and are free to vary under the design algorithm.

Using notation similar to that of~\cite{Girod_ecquant}, the conventional entropy-constrained quantizer design process for an $N$-level quantizer is summarized in Algorithm~\ref{alg:ec_conventional}. With this process, quantizers can be designed to cover a range of rate-distortion performance points, based upon the value of the Lagrange multiplier $\lambda$. To design quantizers that work well in a lightweight compression system for clipped activations, the modified quantizer design process is shown in Algorithm~\ref{alg:ec_mod}. The main modifications are related to pinning the representative levels for the boundary bins and using codeword lengths instead of probabilities for computing rate-related terms. The boundary pinning occurs in Step~\ref{stp:ec_mod_update}. Here, the first and last representative levels are always set to the minimum and maximum activation clipping values ${c}_\mathrm{min}$ and ${c}_\mathrm{max}$, respectively. The interior levels are computed as in Step~\ref{stp:ec_update} of Algorithm~\ref{alg:ec_conventional}.
For modifying the rate terms, we replace the probability-based estimate of the number of bits used to represent a bin, $\log_{2}(p_n)$, with the known length $b_n$ of the binarized codeword used to represent the bin. In the next subsection, we discuss the binarization and subsequent entropy coding.
\begin{algorithm}[!htb]
 \small
\caption{Conventional entropy-constrained quantizer design process}
\label{alg:ec_conventional}
    \hspace*{\algorithmicindent} \textbf{Input:} Training samples \\
    \hspace*{2cm} $x
    \in \{x_m; m = 0,1,\dotsc,M-1\}$ \\
    \hspace*{1.6cm} Number of quantizer bins $N$ \\
    \hspace*{1.6cm} Lagrange multiplier $\lambda$ \\
    \hspace*{\algorithmicindent} \textbf{Output:} Quantizer representative levels \\
    \hspace*{2.2cm} $\hat{x}_{n}, n \in \{0,1, \dotsc ,N-1\}$ \\
    \hspace*{1.85cm} Quantizer decision thresholds \\
    \hspace*{2.2cm} $t_{n}, n \in \{1, \dotsc ,N-1\}$
\begin{algorithmic}[1]
\State  Initialize the representative levels $\hat{x}_n$ and probabilities $p_n$ for each bin, for $n \in \{0,1,\dotsc,N-1\}$ (e.g., uniform and equiprobable)
\State \label{stp:ec_assign}Assign each training sample $x_m$, $m \in \{0,1,\dotsc,M-1\}$ to quantizer bin $n$ having representative value $\hat{x}_n$ such that the Lagrangian rate-distortion cost is minimized:
    \begin{equation*}
        \argmin_{n} \left[ (x_m - \hat{x}_n)^{2} - \lambda \log_{2}p_n \right]
    \end{equation*}
    The subset of samples $x_m$ assigned to bin $n$ is denoted as $\bm{B}_n$.
\State \label{stp:ec_update}Update the probabilities $p_n$ based on the assignment from Step~\ref{stp:ec_assign}, and recompute the representative levels for each bin:
    \begin{equation*}
        \hat{x}_n = \frac{1}{|\bm{B}_n|}\sum_{x \in \bm{B}_n} x \, , \quad n \in \{0,1, \dotsc ,N-1\}
    \end{equation*}
    where $|\bm{B}_n|$ is the number of samples assigned to bin $n$.
\State Based on the recomputed representative levels, recompute the Lagrangian cost function, and repeat Steps~\ref{stp:ec_assign} and~\ref{stp:ec_update} until the reduction in the cost function is less than a threshold.
\State Compute $N-1$ quantizer decision thresholds:
\begin{equation*}
  \begin{split}
    t_{n} = \frac{\hat{x}_n + \hat{x}_{n-1}}{2}
    + \lambda \frac{\log_{2} p_n - \log_{2} p_{n-1}}{2(\hat{x}_n - \hat{x}_{n-1})} \, , \\
    n \in \{1, \dotsc ,N-1\}
  \end{split}
\end{equation*}
\end{algorithmic}
\end{algorithm}
\begin{algorithm}[!htb]
\small
\caption{Modified entropy-constrained quantizer design process for clipped activations}
\label{alg:ec_mod}
    \hspace*{\algorithmicindent} \textbf{Input:} Training samples \\
    \hspace*{2cm} $x
    \in \{x_m; m = 0,1,\dotsc,M-1\}$ \\
    \hspace*{1.6cm} Number of quantizer bins $N$ \\
    \hspace*{1.6cm} Codeword lengths $b_{n}, n \in \{1, \dotsc ,N-1\}$ \\
    \hspace*{1.6cm} Lagrange multiplier $\lambda$ \\
    \hspace*{1.6cm} Activation clipping range $\left[{c}_{\mathrm{min}}, {c}_{\mathrm{max}}\right]$ \\
    \hspace*{\algorithmicindent} \textbf{Output:} Quantizer representative levels \\
    \hspace*{2.2cm} $\hat{x}_{n}, n \in \{0,1, \dotsc ,N-1\}$ \\
    \hspace*{1.85cm} Quantizer decision thresholds \\
    \hspace*{2.2cm} $t_{n}, n \in \{1, \dotsc ,N-1\}$
\begin{algorithmic}[1]
\State Clip (clamp) the training samples $x$ to be within $\left[{c}_{\mathrm{min}}, {c}_{\mathrm{max}}\right]$, which is the clipping range applied to the activations
\State  Initialize the representative levels $\hat{x}_n$ for each bin, for $n \in \{0,1,\dotsc,N-1\}$ (e.g., uniform)
\State \label{stp:ec_mod_assign}Assign each training sample $x_m$, $m \in \{0,1,\dotsc,M-1\}$ to quantizer bin $n$ having representative value $\hat{x}_n$ such that the Lagrangian rate-distortion cost is minimized:
    \begin{equation*}
        \argmin_{n} \left[ (x_m - \hat{x}_n)^{2} - \lambda b_n \right]
    \end{equation*}
    The subset of samples $x_m$ assigned to bin $n$ is denoted as $\bm{B}_n$.
\State \label{stp:ec_mod_update}Recompute the representative levels for each bin:
\begin{align*}
   \hat{x}_0 &= {c}_{\mathrm{min}} \\
   \hat{x}_{N-1} &= {c}_{\mathrm{max}} \\
   \mathrm{if} N > 2:\\
   \hat{x}_n &= \frac{1}{|\bm{B}_n|}\sum_{x \in \bm{B}_n} x \, , \quad n \in \{1, \dotsc ,N-2\}
 \end{align*}
     where $|\bm{B}_n|$ is the number of samples assigned to bin $n$.
\State Based on the recomputed representative levels, recompute the Lagrangian cost function, and repeat Steps~\ref{stp:ec_mod_assign} and~\ref{stp:ec_mod_update} until the reduction in the cost function is less than a threshold.
\State Compute $N-1$ quantizer decision thresholds:
\begin{equation*}
  \begin{split}
    t_{n} = \frac{\hat{x}_n + \hat{x}_{n-1}}{2}
    + \lambda \frac{b_{n} - b_{n-1}}{2(\hat{x}_n - \hat{x}_{n-1})} \, , \\
    n \in \{1, \dotsc ,N-1\}
  \end{split}
\end{equation*}
\end{algorithmic}
\end{algorithm}

\subsection{Binarization and entropy coding}
\begin{figure}[tb]
    \centering
    \includegraphics[width=0.48\textwidth,viewport=6.840000 11.052000 703.107400 195.875432,clip]{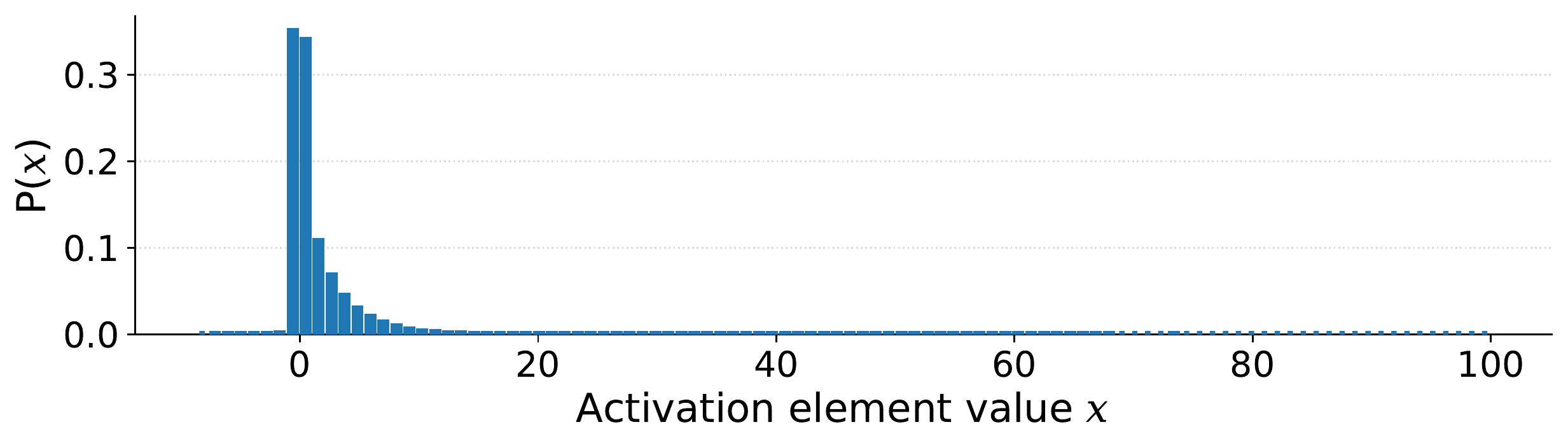}
    \caption{Histogram of activation elements for layer 21 of ResNet-50}
    \label{fig:resnet50_hist}
\end{figure}

After quantizing an activation element, an index associated with the selected representative level is coded and signaled to a bit-stream. For the types of DNNs that we are splitting, the activation values tend to be skewed toward zero, as illustrated by the histogram in Fig.~\ref{fig:resnet50_hist} for the unclipped layer-21 activations of ResNet-50 when run over the ImageNet ILSVRC2012 validation data set.
Given that we will be able to achieve good performance when quantizing to very few bins, a truncated unary binarization scheme~\cite{Marpe2003_CABAC} is well suited for this purpose. For example, the bin indices $\{0,1,2,3\}$ of a 2-bit (4-level) quantizer are mapped to the binarized strings $\{0,10,110,111\}$ respectively.

The binarized strings must next be coded to a bit-stream. For this paper, we use Context-based Adaptive Binary Arithmetic Coding (CABAC)~\cite{Marpe2003_CABAC}, similar to that used in H.264/AVC and HEVC. One context is used for each bit position for the binarized string. For the 2-bit example given in the previous paragraph, three contexts are used.

\section{Experimental results}
\label{sec:experiments}
We applied our lightweight compression technique to activations output from the split layer of a DNN, for three different inference scenarios: YOLOv3 object detection at layer 12, VGG-16 classification at layer 6, and ResNet-50 classification at layer 21. The dimensions of the activations at these layers were 52$\times$52$\times$256, 56$\times$56$\times$128, and 32$\times$32$\times$512, respectively. Pre-trained network weights were obtained from~\cite{darknet_weights}. The software modified to run the experiments was the \textit{Darknet} version of~\cite{AlexeyAB_darknet}. For YOLOv3 with input size 416$\times$416, mAP (IoU~=~0.5) results were obtained using the COCO 2017 validation data set, which includes about 5k images. For VGG-16 and ResNet-50, classification accuracies were obtained using the ImageNet ILSVRC2012 validation data set, which has 50k images. For experiments using entropy-constrained quantization, the quantizer design algorithms were run on activations output when running the first part of the network on 100 images from the data set. After clipping, quantization, and coding to a bit-stream, the activations were decoded and inverse quantized and then passed to the remainder of the neural network. The bit-streams also included side information needed by the decoder, e.g. $c_{\mathrm{min}}$, $c_{\mathrm{max}}$, $N$, and some dimensional parameters for object detection, which together comprised 24 bytes for object detection and 12 bytes for classification networks. The clipping thresholds were empirically selected as described in Section~\ref{subsec:clipping}. The size of the compressed data is reported as bits per element, i.e. the size of the bit-stream divided by the number of elements in the activation's output feature tensor.

The performance when using a uniform quantizer is shown in Tables~\ref{tab:table_yolov3}, \ref{tab:table_vgg16}, and~\ref{tab:table_resnet50} for YOLOv3, VGG-16, and ResNet-50, respectively. For YOLOv3, no loss in performance occurred when quantizing all the way down to 16 levels (4-bit quantization). The drop in mAP was less than 1\% with a 4-level (2-bit) quantizer. For VGG-16, the clipping ranges were much larger than with the other networks, because VGG-16 does not use batch normalization. The Top-1 accuracy drop with an 8-level (3-bit) quantizer was 0.3\% and 1.2\% with 7-level quantization. For ResNet-50, the activations could be quantized to 8 levels with no loss, and at 4 levels (2 bits), the Top-1 loss was well below 1\%. One-bit quantization was feasible with YOLOv3 and ResNet-50, which yielded 4.8\% and 4.9\% losses, respectively, with compressed sizes of 0.39 and 0.41 bits per element. We can also see that as the number of quantization levels decreased, generally the optimal clipping range decreased as well, as discussed in Section~\ref{subsec:clipping}.
\begin{table}[tb]
    \centering
    \caption{Object detection performance for YOLOv3 with uniform quantization and lightweight compression at layer 12}
    \label{tab:table_yolov3}
    \resizebox{\columnwidth}{!}{
    \begin{tabular}{cccccc}
        &
        \makecell{\textbf{quantized} \\ \textbf{bits}} &
        \makecell{\textbf{quantizer} \\ \textbf{bins}} &
        \makecell{\textbf{clip} \\ \textbf{min, max}} &
        \makecell{\textbf{compressed} \\ \textbf{bits/element}} &
        \makecell{\textbf{mAP (\%)}} \\
        \midrule
        \makecell{unmodified \\ (float32)} & 32 & --- & --- & --- & 67.4 \\
        \arrayrulecolor{lightgray} \midrule \arrayrulecolor{defaultrulecolor}
        \multirow{5}{*}{\makecell{with \\ lightweight \\ compression}}
        & 4 & 16 & -0.8, 3.5 & 1.94 & 67.4 \\
        & 3 &  8 & -1.0, 4.0 & 1.15 & 66.8 \\
        & 2 &  4 & -0.1, 2.5 & 0.86 & 66.5 \\
        &   &  3 & -0.1, 2.7 & 0.60 & 66.2 \\
        & 1 &  2 & \phantom{-}0.0, 2.0 & 0.39 & 62.6 \\
    \end{tabular}
    }
 \end{table}
\begin{table}[tb]
    \centering
    \caption{Classifier performance for VGG-16 with uniform quantization and lightweight compression of layer 6 activations}
    \label{tab:table_vgg16}
    \resizebox{\columnwidth}{!}{
    \begin{tabular}{c@{\hskip3pt}c@{\hskip3pt}c@{\hskip3pt}c@{\hskip3pt}c@{\hskip3pt}c@{\hskip3pt}c@{\hskip3pt}}
        &
        \makecell{\textbf{quantized} \\ \textbf{bits}} &
        \makecell{\textbf{quantizer} \\ \textbf{bins}} &
        \makecell{\textbf{clip} \\ \textbf{min, max}} &
        \makecell{\textbf{compressed} \\ \textbf{bits/element}} &
        \makecell{\textbf{Top-1} \\ \textbf{accuracy} \\ \textbf{(\%)}} &
        \makecell{\textbf{Top-5} \\ \textbf{accuracy} \\ \textbf{(\%)}} \\
        \midrule
        \makecell{unmodified \\ (float32)} & 32 & --- & --- & --- & 70.4 & 89.8 \\
        \arrayrulecolor{lightgray} \midrule \arrayrulecolor{defaultrulecolor}
        \multirow{6}{*}{\makecell{with \\ lightweight \\ compression}}
        & 4 & 16 & 0, 2600 & 2.33 & 70.2 & 89.7 \\
        & 3 &  8 & 0, 2400 & 2.41 & 70.1 & 89.6 \\
        &   &  7 & 0, 2200 & 1.57 & 69.2 & 89.1 \\
        &   &  6 & 0, 2000 & 1.48 & 68.6 & 88.7 \\
        &   &  5 & 0, 1800 & 1.36 & 67.8 & 88.2 \\
        & 2 &  4 & 0, 1600 & 1.20 & 66.1 & 87.2 \\
    \end{tabular}
    }
 \end{table}
 \begin{table}[tb]
    \centering
    \caption{Classifier performance for ResNet-50 with uniform quantization and lightweight compression at layer 21}
    \label{tab:table_resnet50}
    \resizebox{\columnwidth}{!}{
    \begin{tabular}{c@{\hskip3pt}c@{\hskip3pt}c@{\hskip3pt}c@{\hskip3pt}c@{\hskip3pt}c@{\hskip3pt}c@{\hskip3pt}}
        &
        \makecell{\textbf{quantized} \\ \textbf{bits}} &
        \makecell{\textbf{quantizer} \\ \textbf{bins}} &
        \makecell{\textbf{clip} \\ \textbf{min, max}} &
        \makecell{\textbf{compressed} \\ \textbf{bits/element}} &
        \makecell{\textbf{Top-1} \\ \textbf{accuracy} \\ \textbf{(\%)}} &
        \makecell{\textbf{Top-5} \\ \textbf{accuracy} \\ \textbf{(\%)}} \\
        \midrule
        \makecell{unmodified \\ (float32)} & 32 & --- & --- & --- & 75.8 & 92.9 \\
        \arrayrulecolor{lightgray} \midrule \arrayrulecolor{defaultrulecolor}
        \multirow{6}{*}{\makecell{with \\ lightweight \\ compression}}
        & 3 &  8 & 0, 14 & 1.23 & 75.8 & 92.9 \\
        &   &  7 & 0, 12 & 1.22 & 75.7 & 92.9 \\
        &   &  6 & 0, 12 & 1.09 & 75.6 & 92.8 \\
        &   &  5 & 0, 12 & 0.94 & 75.5 & 92.8 \\
        & 2 &  4 & 0, 10 & 0.86 & 75.2 & 92.7 \\
        & 1 &  2 & 0, \phantom{1}7 & 0.41 & 70.9 & 90.4 \\
    \end{tabular}
    }
 \end{table}

\begin{figure}[tb]
    \centering
    \includegraphics[width=0.46\textwidth,viewport=7.200000 7.218000 701.099979 421.829987,clip]{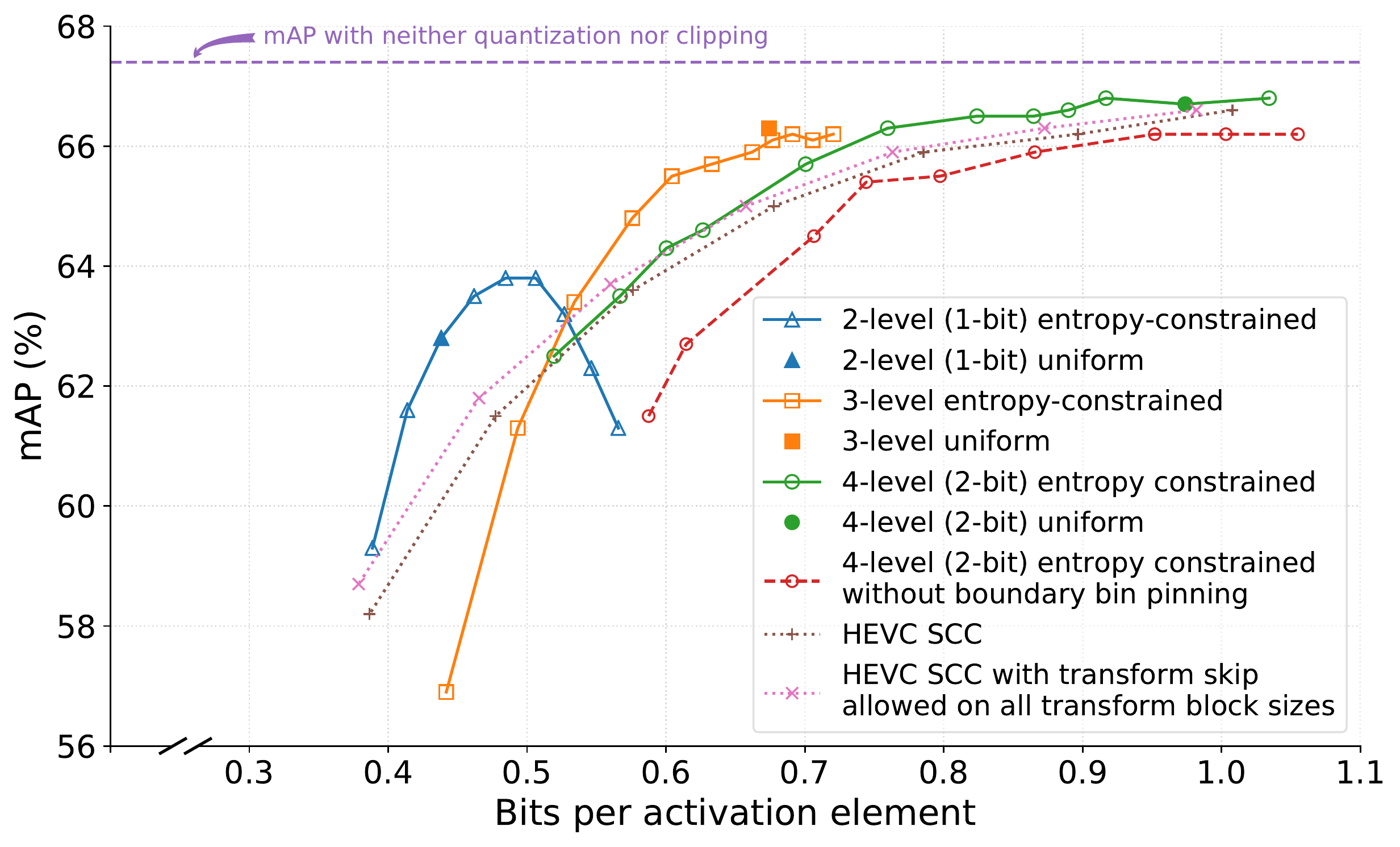}
    \caption{Object-detection mAP for YOLOv3 using lightweight compression at layer 12 with modified entropy-constrained quantization}
    \label{fig:yolov3_plot}
\end{figure}

\begin{figure}[tb]
    \centering
    \includegraphics[width=0.46\textwidth,viewport=7.200000 7.218000 701.063979 421.811987,clip]{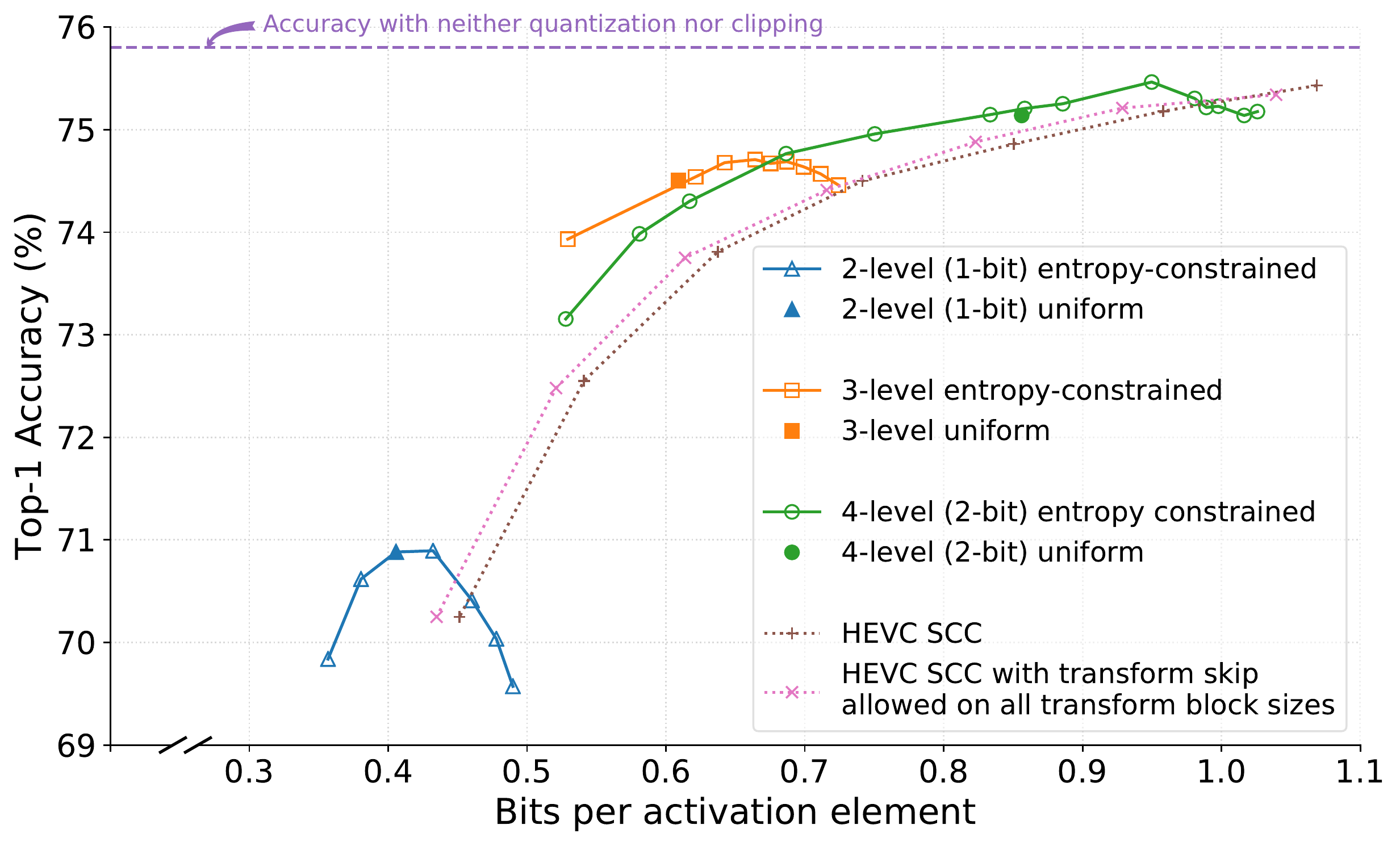}
    \caption{Top-1 accuracy for ResNet-50 using lightweight compression at layer 21 with modified entropy-constrained quantization}
    \label{fig:resnet50_top1_plot}
\end{figure}

\begin{figure}[tb]
    \centering
    \includegraphics[width=0.46\textwidth,viewport=7.200000 7.218000 701.081979 421.847987,clip]{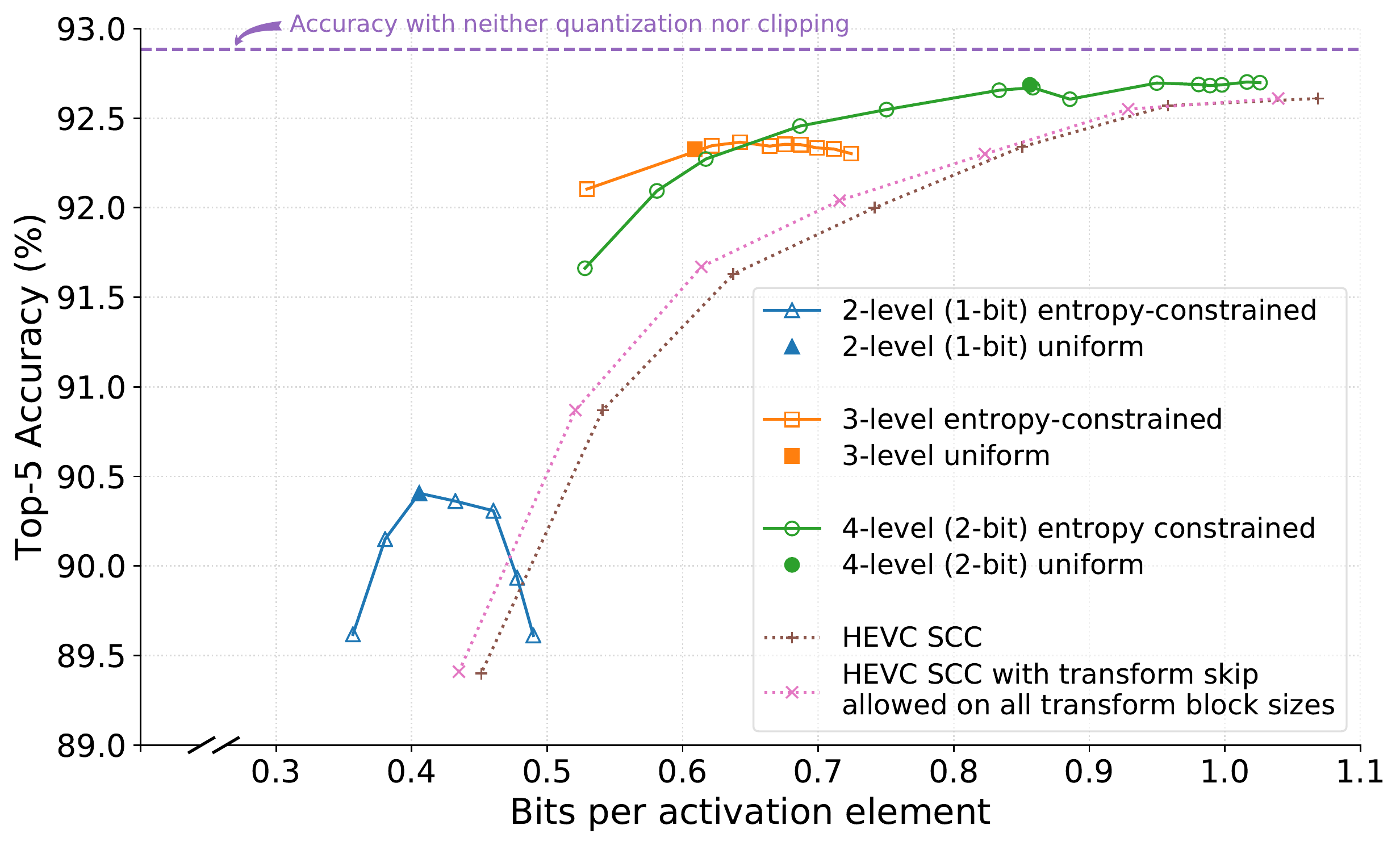}
    \caption{Top-5 accuracy for ResNet-50 using lightweight compression at layer 21 with modified entropy-constrained quantization}
    \label{fig:resnet50_top5_plot}
\end{figure}

Object detection and classification performance when quantizing to between 1 and 2 bits is shown in Figs.~\ref{fig:yolov3_plot}--\ref{fig:resnet50_top5_plot}, for lightweight compression with both uniform and modified entropy-constrained quantization. Fig.~\ref{fig:yolov3_plot} also shows the YOLOv3 performance with a conventional entropy-constrained quantizer that does not pin the outermost reconstruction levels. Here, with 2-bit quantization, the modified method performed about 0.7\% better than with the conventional method, with additional gains at lower rates.
Fig.~\ref{fig:yolov3_plot} also shows that the entropy-constrained quantizer's ability to cover a range of rates enables us to improve the 1-bit performance by about 1\%. For the Top-1 and Top-5 accuracies of ResNet-50, Figs.~\ref{fig:resnet50_top1_plot} and~\ref{fig:resnet50_top5_plot} show that the modified quantizer design algorithm also allows us to obtain improved accuracies for 3- and 4-level (2-bit) quantizers. The relative behaviors among corresponding performance curves for Top-1 and Top-5 accuracies are consistent, with lower losses for Top-5. The range of achievable compressed bit-stream sizes for all these experiments with quantization to 2 and fewer bits was between about 0.3 to 1.0 bits per tensor element.

Figs.~\ref{fig:yolov3_plot}--\ref{fig:resnet50_top5_plot} also show the performance when coding the activations using the HM16.20~\cite{HM16.20} implementation of the HEVC screen content coding extension (HEVC-SCC). HEVC-SCC includes tools that help with the coding of non-camera-captured pictures. As shown in~\cite{Choi2018NearLosslessDF}, when activation channels are arranged to form a picture, they contain much high-frequency content. HEVC-SCC includes a transform skip (TS) mode that is available for all transform block sizes, so we show results when enabling TS for 4$\times$4 blocks only, and for TS enabled over all block sizes. Each set of activation channels were quantized to 8 bits and mosaicked into an 832$\times$832 picture for YOLOv3 and 1024$\times$512 for ResNet-50. Given the fineness of the quantizer, clipping was not necessary. The mosaicked activations for the validation set were coded by HEVC-SCC as an all-Intra sequence of monochrome (4:0:0) 8-bit pictures.
Even with the improved performance with TS on all block sizes, the lightweight compression outperformed HEVC-SCC by up to 1.3\%, depending upon rate. If we compare the operations needed by the lightweight codec to that reported for HEVC~\cite{Bossen1012_hevc_complexity}, we can see that the lightweight codec is well over 90\% less complex than HEVC, as we are only performing clipping, scalar quantization, binarization of only a few symbols, and entropy coding.

\section{Conclusions}
\label{sec:conclusions}
We presented an efficient post-training lightweight compression method for collaborative intelligence applications, suitable for when an edge device compresses the output of a split DNN and transmits it to the cloud, where the remainder of the DNN is processed.
The codec only requires clipping, coarse quantization, binarization, and entropy coding to compress feature tensors, so is well over 90\% less complex than existing image or video codecs such as HEVC that are typically used for picture compression. We also presented an entropy-constrained quantizer design process tailored specifically for clipped activations. With this lightweight lossy coding technique, we were able to quantize the 32-bit floating point activations at a split DNN layer to fewer than 2 bits per element and then compress them further to 0.6 to 0.8 bits per element, while keeping the network accuracy loss to less than 1\%. When coding the activations using HEVC, we showed that the lightweight codec yielded accuracies of up to 1.3\% higher than HEVC, depending upon the rate. The performance and simplicity of this lightweight compression technique makes it an attractive option for coding activations for edge/cloud DNN applications.

\bibliographystyle{IEEEbib-abbrev}
\bibliography{IEEEabrv,strings,refs}

\begin{thebibliography}{10}

\bibitem{Chen2019}
J. {Chen} and X. {Ran},
\newblock ``Deep learning with edge computing: A review,''
\newblock {\em Proceedings of the IEEE}, vol. 107, no. 8, pp. 1655--1674, Aug.
  2019.

\bibitem{Lane2018}
N.~D. {Lane} and P. {Warden},
\newblock ``The deep (learning) transformation of mobile and embedded
  computing,''
\newblock {\em Computer}, vol. 51, no. 5, pp. 12--16, May 2018.

\bibitem{Tan_2019_CVPR}
M. Tan, B. Chen, R. Pang, V. Vasudevan, and Q.~V. Le,
\newblock ``Mnasnet: Platform-aware neural architecture search for mobile,''
\newblock {\em 2019 IEEE/CVF Conference on Computer Vision and Pattern
  Recognition (CVPR)}, pp. 2815--2823, June 2018.

\bibitem{eshratifar2019towards}
A.~E. Eshratifar, A. Esmaili, and M. Pedram,
\newblock ``Towards collaborative intelligence friendly architectures for deep
  learning,''
\newblock {\em 20th Int. Symposium Quality Electronic Design (ISQED)}, pp.
  14--19, Mar. 2019.

\bibitem{Kang2017}
Y. Kang, J. Hauswald, C. Gao, A. Rovinski, T.~N. Mudge, J. Mars, and L. Tang,
\newblock ``Neurosurgeon: Collaborative intelligence between the cloud and
  mobile edge,''
\newblock in {\em ASPLOS '17}, Apr. 2017.

\bibitem{Mishra2017_WRPN}
A. Mishra, E. Nurvitadhi, J.~J. Cook, and D. Marr,
\newblock ``{WRPN}: Wide reduced-precision networks,''
\newblock in {\em 6th Int. Conf. on Learning Representations (ICLR)}, May 2018.

\bibitem{Banner2018_8BitTraining}
R. Banner, I. Hubara, E. Hoffer, and D. Soudry,
\newblock ``Scalable methods for 8-bit training of neural networks,''
\newblock in {\em Proc. 32nd Int. Conf. Neural Information Processing Systems
  ({NeurIPS})}, Dec. 2018, pp. 5151--–5159.

\bibitem{Choi2019_2bit}
J. Choi, S. Venkataramani, V. Srinivasan, K. Gopalakrishnan, Z. Wang, and P.
  Chuang,
\newblock ``Accurate and efficient 2-bit quantized neural networks,''
\newblock in {\em Proc. 2\textsuperscript{nd} {SysML} Conf.}, Mar. 2019.

\bibitem{Hubara2016_BNN}
I. Hubara, M. Courbariaux, D. Soudry, R. El-Yaniv, and Y. Bengio,
\newblock ``Binarized neural networks,''
\newblock in {\em Proc. 30th Int. Conf. Neural Information Processing Systems
  ({NIPS})}, Dec. 2016, pp. 4114--–4122.

\bibitem{Rastegari2016_XNOR}
M. Rastegari, V. Ordonez, J. Redmon, and A. Farhadi,
\newblock ``{XNOR-Net}: {ImageNet} classification using binary convolutional
  neural networks,''
\newblock in {\em 14th European Conf. on Computer Vision ({ECCV})}, Oct. 2016,
  pp. 525--542.

\bibitem{Zhao2019_Quantization}
R. Zhao, Y. Hu, J. Dotzel, C.~D. Sa, and Z. Zhang,
\newblock ``Improving neural network quantization without retraining using
  outlier channel splitting,''
\newblock in {\em Proc. 36th Int. Conf. on Machine Learning, {ICML} 2019}, June
  2019, pp. 7543--7552.

\bibitem{banner2018_ACIQ}
R. Banner, Y. Nahshan, E. Hoffer, and D. Soudry,
\newblock ``{ACIQ}: Analytical clipping for integer quantization of neural
  networks,'' [Online]: \url{https://openreview.net/forum?id=B1x33sC9KQ}, Sept.
  2018.

\bibitem{Banner2019_4bit}
R. Banner, Y. Nahshan, E. Hoffer, and D. Soudry,
\newblock ``Post-training 4-bit quantization of convolution networks for
  rapid-deployment,''
\newblock in {\em Proc. 33rd Int. Conf. Neural Information Processing Systems
  ({NeurIPS})}, May 2019, pp. 7950--7958.

\bibitem{Krishnamoorthi2018_QuantizingDC}
R. Krishnamoorthi,
\newblock ``Quantizing deep convolutional networks for efficient inference: A
  whitepaper,''
\newblock {\em arXiv abs/1806.08342}, June 2018.

\bibitem{dfc_for_collab_object_detection}
H. Choi and I.~V. Baji\'{c},
\newblock ``Deep feature compression for collaborative object detection,''
\newblock {\em Proc. 25th IEEE Int. Conf. Image Processing (ICIP)}, pp.
  3743--3747, Oct. 2018.

\bibitem{Choi2018NearLosslessDF}
H. Choi and I.~V. Baji\'{c},
\newblock ``Near-lossless deep feature compression for collaborative
  intelligence,''
\newblock {\em IEEE 20th Int. Workshop on Multimedia Signal Processing (MMSP)},
  Aug. 2018.

\bibitem{eshratifar2019bottlenet}
A.~E. {Eshratifar}, A. {Esmaili}, and M. {Pedram},
\newblock ``{BottleNet}: A deep learning architecture for intelligent mobile
  cloud computing services,''
\newblock in {\em Proc. IEEE/ACM Int. Symposium Low Power Electronics and
  Design (ISLPED)}, July 2019.

\bibitem{Choi_BaF_2020}
H. Choi, R.~A. Cohen, and I.~V. Baji\'{c},
\newblock ``Back-and-forth prediction for deep tensor compression,''
\newblock {\em Proc. 45th IEEE Int. Conf. Acoustics, Speech, and Signal
  Processing (ICASSP)}, May 2020,
\newblock in press.

\bibitem{hevc_scc}
J. Xu, R. Joshi, and R.~A. Cohen,
\newblock ``Overview of the emerging {HEVC} screen content coding extension,''
\newblock {\em {IEEE} Trans. Circuits Syst. Video Technol.}, vol. 26, no. 1,
  pp. 50--62, Jan. 2016.

\bibitem{Vanhoucke2011}
V. Vanhoucke, A. Senior, and M.~Z. Mao,
\newblock ``Improving the speed of neural networks on {CPUs},''
\newblock in {\em Deep Learning and Unsupervised Feature Learning Workshop,
  NIPS}, Dec. 2011.

\bibitem{He2015DeepRL}
K. He, X. Zhang, S. Ren, and J. Sun,
\newblock ``Deep residual learning for image recognition,''
\newblock {\em {IEEE} Conf. Computer Vision and Pattern Recognition ({CVPR})},
  pp. 770--778, June 2016.

\bibitem{imagenet2015}
O. Russakovsky et~al.,
\newblock ``{ImageNet Large Scale Visual Recognition Challenge},''
\newblock {\em Int. J. Comput. Vision}, vol. 115, no. 3, pp. 211--252, Dec.
  2015.

\bibitem{Redmon2018_yolov3}
J. Redmon and A. Farhadi,
\newblock ``{YOLOv3}: An incremental improvement,''
\newblock {\em arXiv preprint arXiv:1804.02767}, Apr. 2018.

\bibitem{COCO}
T.-Y. Lin, M. Maire, S. Belongie, J. Hays, P. Perona, D. Ramanan, P.
  Doll{\'a}r, and C.~L. Zitnick,
\newblock ``Microsoft {COCO}: Common objects in context,''
\newblock in {\em European Conf. on Computer Vision (ECCV)}, Sept. 2014.

\bibitem{Chou1989}
P.~A. {Chou}, T. {Lookabaugh}, and R.~M. {Gray},
\newblock ``Entropy-constrained vector quantization,''
\newblock {\em {IEEE} Trans. Acoust., Speech, Signal Process.}, vol. 37, no. 1,
  pp. 31--42, Jan. 1989.

\bibitem{Sullivan1998}
G.~J. {Sullivan} and T. {Wiegand},
\newblock ``Rate-distortion optimization for video compression,''
\newblock {\em IEEE Signal Processing Magazine}, vol. 15, no. 6, pp. 74--90,
  Nov. 1998.

\bibitem{Girod_ecquant}
B. Girod,
\newblock ``Quantization,'' {EE398A} Image and Video Compression, [Online]:
  \url{https://web.stanford.edu/class/ee398a/handouts/lectures/05-Quantization.pdf}.

\bibitem{Marpe2003_CABAC}
D. Marpe, H. Schwarz, and T. Wiegand,
\newblock ``Context-based adaptive binary arithmetic coding in the {H.264/AVC}
  video compression standard,''
\newblock {\em {IEEE} Trans. Circuits Syst. Video Technol.}, vol. 13, no. 7,
  pp. 620--636, July 2003.

\bibitem{darknet_weights}
J. Redmon,
\newblock ``Darknet: {O}pen source neural networks in {C},'' [Online]:
  \url{https://pjreddie.com/darknet}.

\bibitem{AlexeyAB_darknet}
A. Bochkovskiy,
\newblock ``{darknet},'' [Online]:
  \url{https://github.com/AlexeyAB/darknet/tree/8c80ba6}.

\bibitem{HM16.20}
``{HEVC} reference software ({HM} 16.20),''
  \url{http://hevc.hhi.fraunhofer.de/svn/svn_HEVCSoftware/tags/HM-16.20+SCM-8.8},
\newblock Accessed: 2019-12-12.

\bibitem{Bossen1012_hevc_complexity}
F. Bossen, B. Bross, K. Suhring, and D. Flynn,
\newblock ``{HEVC} complexity and implementation analysis,''
\newblock {\em IEEE Trans. Circuits Syst. Video Technol.}, vol. 22, no. 12, pp.
  1685--1696, Dec. 2012.

\end{thebibliography}

\end{document}